\documentclass[conference]{IEEEtran}
\IEEEoverridecommandlockouts
\usepackage{cite}
\usepackage{amsmath,amssymb,amsfonts}
\usepackage{tabularx}
\usepackage{booktabs}
\usepackage{bm}
\usepackage{hyperref}
\usepackage{multirow}
\usepackage{enumitem}
\usepackage{subcaption}
\usepackage{textcomp}
\usepackage{graphicx}
\usepackage{epsfig}
\usepackage{textcomp}
\usepackage{xcolor}
\usepackage{mathrsfs}
\usepackage{eucal}
\usepackage{pifont}
\usepackage{amsthm}
\usepackage{float}
\usepackage{color}
\usepackage{dsfont}
\usepackage{algorithm}
\usepackage{algpseudocode}

\newtheorem{remark}{Remark}

\def\BibTeX{{\rm B\kern-.05em{\sc i\kern-.025em b}\kern-.08em
    T\kern-.1667em\lower.7ex\hbox{E}\kern-.125emX}}
\begin{document}
\title{\vspace{7.8mm} \LARGE \bf ProMix: Efficient Data Mixture Optimization for Robotic Imitation via Proxy-Reference DRO}

\author{\IEEEauthorblockN{Qie Sima$^{*}$, Wei Xue, Yike Guo$^{\dagger}$}
\IEEEauthorblockA{
\textit{Hong Kong University of Science and Technology}
}
}

\maketitle

\begin{abstract}
Vision-Language-Action (VLA) models excel in open-world manipulation, yet their performance often saturates when training data are naively pooled across heterogeneous robots, tasks, and environments. Distributionally robust data mixture optimization, exemplified by Re-Mix, can recover much of this performance, but its weight search requires training full-scale policies twice on the entire corpus, placing it out of reach for most laboratories. We present ProMix, a compute-efficient instantiation of the proxy-reference DRO recipe for robotic imitation learning: a 15M-parameter proxy learns per-domain sampling weights by minimizing worst-case excess loss against a frozen reference model, and the learned weights transfer, without re-tuning, to VLA backbones up to 7B parameters. We further provide, to our knowledge, the first controlled analysis of the reference model's role in robotic DRO, showing that removing it induces over-pessimism toward noisy, low-quality domains ($-3.8\%$ average and $-14.3\%$ worst-domain success rate), and that entropy regularization is needed to prevent weight collapse. Evaluated on 9500 real-world episodes from the Open-X Embodiment dataset, ProMix improves the average success rate over uniform mixing from 68.4\% to 76.2\%, matching Re-Mix within evaluation noise while reducing mixture-optimization cost to a backbone-independent 11 GPU-hours---versus 16 GPU-hours at the 50M comparison scale and 192 GPU-hours at 7B for full-scale searches---with a $+17.1\%$ absolute gain on the worst-case domain.
\end{abstract}

\section{Introduction}
Vision-Language-Action (VLA) models are rapidly transforming the robotics landscape, enabling agents to execute complex, open-world tasks from natural language instructions. The impressive generalization capabilities of these foundation models, such as RT-1 \cite{RT1}, are a direct result of pre-training on massive, diverse multimodal datasets. Recently, the field has gravitated toward ``general-purpose physical intelligence'' with the emergence of the $\pi$ model family \cite{black2024pi_0, intelligence2025pi_}, which achieves state-of-the-art control through sophisticated data curation and distillation. Beyond proprietary efforts, open datasets such as Open-X \cite{openx202} and DROID \cite{khazatsky2024droid} and open models such as OpenVLA \cite{kim2025openvla} and GR00T N1 \cite{bjorck2025gr00t} have made large-scale pre-training broadly accessible. However, a critical bottleneck is shifting from architecture to data: effectively blending heterogeneous data collected from disparate robots, platforms, and tasks. Naively pooling these domains causes performance saturation, and the curation recipes used by industrial labs are neither public nor reproducible.

\begin{figure*}[tb]
    \centering
    \includegraphics[width=0.82\textwidth]{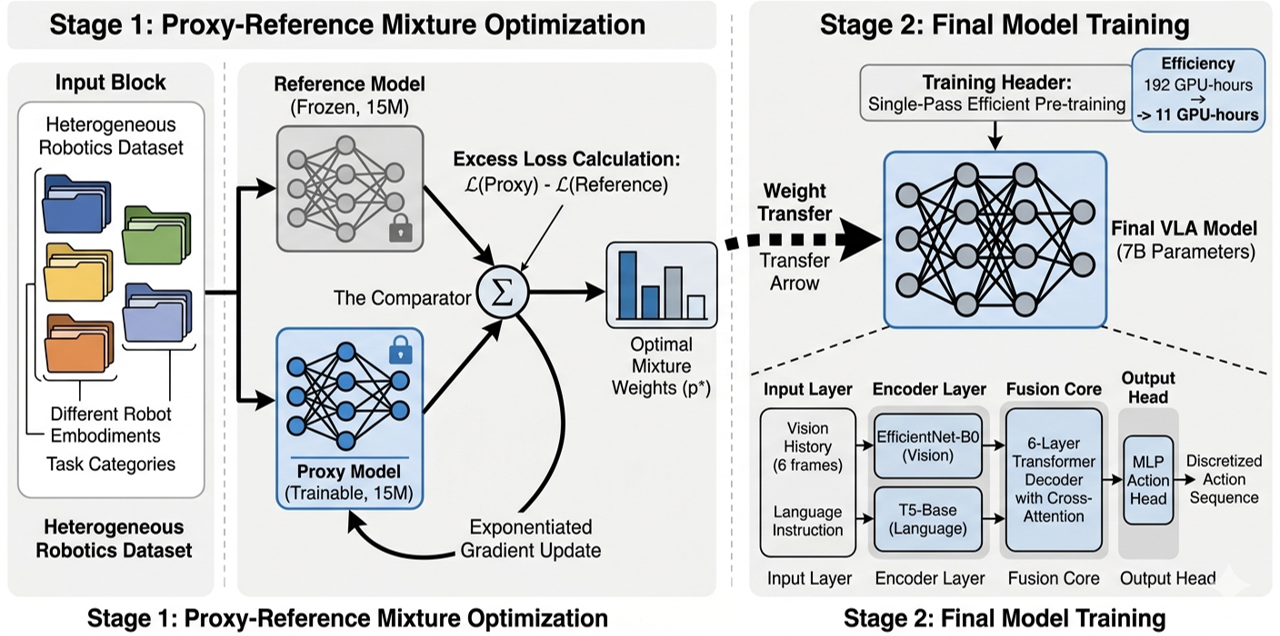}
    \caption{\textbf{The ProMix Framework Architecture.} The pipeline consists of two stages: (Stage 1) a proxy-reference optimization loop that learns optimal domain weights $\mathbf{p}^*$ by minimizing the worst-case excess loss of a 15M trainable proxy against a frozen reference model; (Stage 2) single-pass training of the final VLA model using the transferred weights $\mathbf{p}^*$. The entire mixture search costs 11 GPU-hours,whereas a full-scale Re-Mix search costs 16 GPU-hours at the 50M comparison scale and 192 GPU-hours at 7B.}
    \label{fig:framework}
\end{figure*}
Recent research has started to bridge LLM data optimization and robotics. DoReMi \cite{xie2023doremi} showed that in LLM pre-training, domain weights can be learned cheaply by a small proxy model trained against a frozen reference model, so that the expensive weight search never touches the final large model. Re-Mix \cite{remix} ported this distributionally robust optimization (DRO) objective to large-scale imitation learning and demonstrated substantial gains over uniform and hand-tuned mixtures on Open-X Embodiment. However, Re-Mix's instantiation retains a prohibitive cost structure: both the reference model and the robust proxy are trained on the full corpus at (near-)final policy scale, so the weight search roughly doubles the training budget before the final policy is even trained---a limitation the authors themselves highlight. Moreover, two questions that determine whether such methods are practical at the scale of modern VLA training remain unquantified. First, is the reference model actually load-bearing in robotics, where continuous action spaces and heteroscedastic demonstration noise dominate the loss? Without it, a standalone proxy may become \textbf{overly pessimistic}, disproportionately assigning weights to inherently chaotic or noisy trajectories rather than to domains with the most headroom for cross-domain generalization. Second, do weights learned at a small proxy scale transfer to the large backbones where the savings matter most? This transfer is asserted in prior work but has not been systematically measured for robotic policies.

This paper introduces ProMix, a framework that makes proxy-reference data mixture optimization practical for robotic imitation learning. ProMix learns mixture weights with a 15M-parameter proxy guided by a frozen, standardly-trained reference model, at a cost of 11 GPU-hours that is fixed regardless of the final backbone---under matched accounting, a full-scale search in our reimplementation of Re-Mix costs 16 GPU-hours at the 50M comparison scale and 192 GPU-hours at the 7B scale---and transfers the learned weights, without re-tuning, to 50M and 7B VLA backbones. Beyond efficiency, we use ProMix as an experimental probe of the proxy-reference design itself: we provide, to our knowledge, the first controlled ablation of the reference model in robotic DRO, showing that removing it induces over-pessimism toward noisy domains ($-3.8\%$ average and $-14.3\%$ worst-domain success rate), and that an entropy regularizer on the weight simplex is essential to prevent weight collapse. This offers a practical path for researchers to achieve competitive performance without industrial-scale computing clusters, aligning with the industry trend toward compact yet capable models like $\pi_{0.5}$ \cite{intelligence2025pi_}.

Our key contributions are as follows:
\begin{itemize}
    \item We present ProMix, a compute-efficient proxy-reference framework for data mixture optimization in robotic imitation learning, which reduces mixture-optimization cost to a backbone-independent 11 GPU-hours---versus 16 GPU-hours at 50M and 192 GPU-hours at 7B ($17\times$) for full-scale searches---by performing the entire search with a 15M proxy rather than full-scale policies.
    \item We provide, to our knowledge, the first controlled analysis of the reference model in robotic DRO, demonstrating that the excess-loss objective is essential for mitigating over-pessimism on noisy, heterogeneous robotic data, and we identify entropy regularization as the stabilizer that prevents weight collapse.
    \item We quantify cross-scale transferability on the Open-X Embodiment dataset: weights learned by the 15M proxy consistently improve 50M and 7B VLA models, lifting the average success rate from 68.4\% to 76.2\% and the worst-domain rate from 51.4\% to 68.5\%.
\end{itemize}

\section{Related Work}
\label{sec:related work}

\subsection{Robotics Foundation Models and Data Efficiency}
Foundation models have enabled general-purpose agents to acquire complex skills from large-scale datasets. RT-1 \cite{RT1} demonstrated the potential of transformer-backed Visual-Language-Action (VLA) architectures, and the $\pi$ model family (e.g., $\pi_0$ \cite{black2024pi_0} and $\pi_{0.5}$ \cite{intelligence2025pi_}) shows that sophisticated data curation yields capable policies with a fraction of the parameters. Open models such as OpenVLA \cite{kim2025openvla} and GR00T N1 \cite{bjorck2025gr00t}, together with open corpora such as Open-X \cite{openx202} and DROID \cite{khazatsky2024droid}, have made large-scale pre-training broadly accessible.

The success of these models remains tied to how well heterogeneous data sources are combined. Beyond large teleoperation corpora such as BridgeData \cite{walke2023bridgedata}, existing data-efficiency efforts---synthetic data generation, distillation, and demonstration filtering---typically improve training efficacy while still treating all real-world data equally, overlooking the core challenge: how to \textit{optimally combine} heterogeneous data from disparate sources. ProMix addresses this by automating the discovery of the most informative data blend.

\subsection{Distributionally Robust Learning}
Alternative approaches select training data online---via curriculum sequencing or active human-in-the-loop queries---but require interaction or iterative labeling that does not scale to large-scale pre-training.

Theoretically, our approach is grounded in Distributionally Robust Optimization (DRO), which minimizes worst-case loss over an uncertainty set to guard against distribution shift, here applied in its group-level form over predefined training domains. Recent work on distributionally robust imitation learning \cite{tasil_2025} has highlighted the necessity of DRO to handle data heterogeneity-induced shifts. ProMix operationalizes these insights through a scalable proxy-reference framework. In noisy robotics environments, the reference model converts the objective from absolute worst-case loss to an \emph{excess loss}, which we show empirically is the difference between robust optimization and over-pessimistic weight collapse.

\subsection{Data Pruning and Demonstration-Level Curation}
To handle the scale of robotics datasets, researchers have explored selection and pruning to improve efficiency. At the demonstration level, DemInf \cite{hejna2025deminf} ranks individual trajectories by mutual-information estimates and achieves consistent gains, while quality filtering and subsetting have been used to reduce training costs. Such methods may, however, discard data from minority domains or outlier tasks, potentially harming out-of-distribution robustness. Domain-level mixing and demonstration-level pruning are orthogonal: ProMix retains the entire dataset but dynamically re-weights domains, ensuring that high-variance minority domains contribute appropriately to generalization, and is complementary to trajectory-level curation.

\subsection{Data Mixture Optimization (DMO)}
The concept of learning an optimal data mixture (DMO) originated in LLM pre-training. DoReMi \cite{xie2023doremi} introduced the canonical recipe: a lightweight proxy model is trained with group DRO against a frozen reference model trained on a uniform mixture, minimizing worst-case \emph{excess} loss, and the resulting weights transfer to much larger models. Follow-up LLM work refined this recipe: Data Mixing Laws \cite{ye2025data} fits parametric scaling laws that predict the loss of unseen mixtures, and RegMix \cite{liu2025regmix} fits a regression over small-scale runs to predict large-model performance under unseen mixtures. These alternatives require training multiple candidate mixtures to fit the mixture--performance model; ProMix instead requires a single proxy--reference pair whose search cost is independent of the final backbone.

In robotics, Re-Mix \cite{remix} successfully ported DoReMi's excess-loss DRO to imitation learning, showing large gains over uniform and human-selected weights on Open-X. However, its instantiation leaves two gaps that matter for VLA-scale training. First, the reference model and the robust proxy are trained on the full corpus at (near)final policy scale, so computing the weights roughly doubles the total training budget -- the authors themselves flag this two-pass cost as a key limitation and call for cheaper alternatives. Second, the paper asserts but does not measure transfer of weights across model scales, and provides no ablation isolating the reference model's contribution in the robotic setting. ProMix closes both gaps: it performs the entire search with a 15M proxy and frozen reference (11 GPU-hours, versus 16--192 GPU-hours for full-scale searches at 50M--7B under matched conventions), provides to our knowledge the first controlled study of the reference model's role in robotic DRO, and quantifies cross-scale transfer up to a 7B backbone. We additionally identify entropy regularization on the weight simplex as a practical necessity---without it, the weights collapse onto a few noisy domains, a failure mode we characterize empirically.

\section{Problem Formulation}
\label{sec:problem}

Our objective is to train a large-scale Vision-Language-Action (VLA) model, denoted as $\bm{\theta}$, on a heterogeneous dataset $\mathcal{D}$. The dataset is composed of $N$ distinct data domains, such as different robot embodiments or task categories, represented as $\mathcal{D} = \{\mathcal{D}_1, \mathcal{D}_2, \ldots, \mathcal{D}_N\}$. Each domain $\mathcal{D}_i$ consists of a set of demonstrations $(s, a, l)$, where $s$ is the state (image), $a$ is the action, and $l$ is the language instruction.

In standard imitation learning, the model is trained by minimizing a re-weighted empirical risk over the entire dataset:
\begin{equation}
\min_{\bm{\theta}} \sum_{i=1}^{N} \alpha_i L_{\mathcal{D}_i}(\bm{\theta})
\label{eq:uniform}
\end{equation}
where $L_{\mathcal{D}_i}(\bm{\theta}) = \frac{1}{|\mathcal{D}_i|} \sum_{(s, a, l) \in \mathcal{D}_i} \mathcal{L}(\bm{\theta}; s, a, l)$ is the average behavior cloning loss for domain $\mathcal{D}_i$, and the mixture coefficients $\alpha_i$ are set either uniformly ($\alpha_i = 1/N$) or proportionally to the domain size ($\alpha_i \propto |\mathcal{D}_i|$); we evaluate both choices as baselines in Section~\ref{sec:exp}. However, this approach often leads to performance saturation, as domains with massive but low-quality demonstrations may dominate the gradient, causing the model to underperform on smaller or more challenging "long-tail" domains.

To achieve robustness across heterogeneous sources, we frame data mixture optimization as a \textbf{Distributionally Robust Optimization (DRO)} problem. Instead of minimizing the average loss, DRO aims to minimize the \textit{worst-case} loss across all possible domain mixtures; in Eq.~\eqref{eq:excess_dro} below, the uncertainty set is the simplex over the $N$ training domains (group DRO), so the adversary re-weights whole domains rather than perturbing individual samples, which keeps the inner maximization tractable at dataset scale. However, a naive DRO objective (e.g., $\min_{\bm{\theta}} \max_{\bm{p}} \sum p_i L_{\mathcal{D}_i}(\bm{\theta})$) in robotics often fails because it over-prioritizes domains that are \textit{inherently difficult} or noisy, rather than those where the model has the most room for improvement; we quantify this failure mode in Section~\ref{subsec:results}.

To address this, we define the optimization objective using \textbf{excess loss} relative to a reference model $\bm{\theta}_{ref}$. The goal is to find sampling weights $\bm{p} = [p_1, \ldots, p_N]$ that minimize the worst-case regret across domains:
\begin{equation}
\min_{\bm{\theta}} \max_{\bm{p} \in \Delta} \sum_{i=1}^{N} p_i \cdot \left( L_{\mathcal{D}_i}(\bm{\theta}) - L_{\mathcal{D}_i}(\bm{\theta}_{ref}) \right)
\label{eq:excess_dro}
\end{equation}
where $\Delta = \{\bm{p} \in \mathbb{R}^N \mid \sum_{i=1}^N p_i = 1, p_i \geq 0\}$ is the probability simplex. The term $L_{\mathcal{D}_i}(\bm{\theta}_{ref})$ serves as a baseline for the "achievable" performance on domain $\mathcal{D}_i$. By minimizing the difference, the model focuses on domains where the current policy significantly lags behind the reference, effectively mitigating over-pessimism toward inherently noisy data.

\begin{remark}[Excess loss redirects the worst case toward learnable domains]
\label{rem:toy}
Consider two domains. Domain $A$ is dominated by irreducible demonstration noise: no policy can push its loss below $0.90$, and both the reference and the current proxy sit at this floor, $L_A(\bm{\theta}_R) = L_A(\bm{\theta}_P) = 0.90$. Domain $B$ is learnable: the reference achieves $L_B(\bm{\theta}_R) = 0.50$, the proxy currently sits at $L_B(\bm{\theta}_P) = 0.60$, and a loss of $0.20$ is attainable. Absolute-worst-case weighting assigns all mass to $A$ ($0.90 > 0.60$)---a domain on which no model can improve---and starves $B$ of gradient. Excess-loss weighting assigns all mass to $B$ (excess $+0.10$ on $B$ versus $0$ on $A$), i.e., to the domain with genuine headroom. High \emph{absolute} loss therefore signals noise rather than headroom; only the excess loss separates the two, which is the failure mode we quantify in Section~\ref{subsec:results}.
\end{remark}

Solving Eq.~\eqref{eq:excess_dro} directly for a large VLA model is computationally prohibitive. Our proposed \textbf{ProMix} framework decouples this process by using a lightweight proxy model $\bm{\theta}_{proxy}$ to learn the optimal weights $\bm{p}^*$, which are then transferred to the final large-scale model.
\section{Methodology}
\label{sec:method}

Our proposed \textbf{ProMix} framework introduces a two-stage training pipeline designed to optimize data efficiency in large-scale robotic learning. The first stage, which requires minimal computation, learns the optimal data mixture weights using a lightweight proxy-reference architecture. The second stage uses these learned weights to efficiently train the final, large-scale VLA model. This decoupling is motivated by observations in LLM pre-training \cite{xie2023doremi} that relative domain difficulties identified by compact models often remain consistent when scaling to larger architectures; Section~\ref{sec:exp} tests this assumption directly for robotic imitation.

\subsection{The Proxy-Reference Framework}
\label{subsec:proxy_ref}

Central to our approach is the \textbf{Proxy-Reference Framework}, designed to efficiently learn optimal data mixture weights without the prohibitive cost of training large VLA models repeatedly. This framework operates on the principle that the performance of a small, lightweight model serves as a reliable proxy for a much larger model, especially when evaluating the \textit{relative} learning progress across heterogeneous data domains. We define three core components: a Reference Model ($\bm{\theta}_R$), a Proxy Model ($\bm{\theta}_P$), and a Final Model ($\bm{\theta}_L$).

\subsubsection{VLA Model Architecture}
\label{subsubsec:architecture}

Both the \textbf{Reference Model} and the \textbf{Proxy Model} utilize a compact, shared VLA architecture with approximately 15M \emph{trainable} parameters (the frozen encoders are excluded from this count) optimized for processing high-frequency robotics data. For a given demonstration $(s_{1:T}, a_{1:T}, l)$, the model predicts the next action sequence. The architecture comprises:
\begin{enumerate}
    \item \textbf{Vision Encoder}: A pre-trained EfficientNet-B0 processes a history of $k=6$ images $(s_{t-k}, \ldots, s_t)$ to extract visual features, which are projected to a dimensionality of $d_{\text{model}}=384$.
    \item \textbf{Language Encoder}: A frozen, pre-trained T5-Base model encodes the natural language instruction $l$ into a set of language tokens.
    \item \textbf{Transformer Decoder}: A 6-layer cross-attention decoder processes the visual and language tokens. It attends to both visual features and language tokens to produce a fused, multi-modal representation. An action head predicts the action sequence $(a_{t+1}, \ldots, a_{t+m})$ for the next $m$ timesteps, treating the continuous action space via discretization to handle multi-modality.
\end{enumerate}

\subsubsection{Reference Model}
\label{subsubsec:ref_model}

The \textbf{Reference Model} ($\bm{\theta}_R$) serves as a stable performance benchmark. Trained once on the \textbf{entire dataset $\mathcal{D}$ using a uniform data mixture}, its parameters are subsequently frozen. Its primary role is to provide a baseline for the "inherent difficulty" of each domain $\mathcal{D}_i$. In robotic imitation, certain domains (e.g., noisy human demonstrations) exhibit naturally higher losses. By using $\bm{\theta}_R$ to calculate excess loss, we prevent the optimization from over-prioritizing these "unlearnable" noisy domains, a critical distinction from standard DRO that ensures the proxy model focuses on domains with the highest potential for improvement.

The reference checkpoint is taken after a fixed budget of 100{,}000 optimization steps. The same checkpoint-selection rule is used for every seed, and the reference remains frozen throughout proxy training.

\subsubsection{Proxy Model}
\label{subsubsec:proxy_model}

The \textbf{Proxy Model} ($\bm{\theta}_P$) is the core of our optimization. It shares an identical architecture with $\bm{\theta}_R$ but begins with a fresh initialization to learn the optimal data mixture weights $\bm{p}$. Training uses per-domain mini-batches on an NVIDIA H100 GPU: each iteration samples one batch of 128 from every domain ($N \times 128$ samples per iteration), so that each per-domain excess loss is estimated on an independent, equally sized batch, and the mixture weights enter only through $p_i$ in the loss and in the weight update.
In each training iteration:
\begin{enumerate}
    \item \textbf{Forward Pass}: Each per-domain batch $\mathcal{B}_i$ is passed through $\bm{\theta}_P$ to compute $\ell_i^P=\mathcal{L}(\bm{\theta}_P; \mathcal{B}_i)$ and through the \textbf{frozen} $\bm{\theta}_R$ to compute $\ell_i^R=\mathcal{L}(\bm{\theta}_R; \mathcal{B}_i)$, detached from the gradient graph.
    \item \textbf{Excess Loss Calculation}: The per-domain excess loss is computed as $L_{\mathcal{D}_i}(\bm{\theta}_P) - L_{\mathcal{D}_i}(\bm{\theta}_R)$, representing the current "regret" relative to the baseline.
    \item \textbf{Weight Update}: The domain weights $\bm{p}$ are updated via exponentiated gradient ascent with a learning rate of $\eta=0.01$, as detailed in Section \ref{subsec:learning_weights}.
    \item \textbf{Model Parameter Update}: The proxy model's parameters $\bm{\theta}_P$ are updated via AdamW (learning rate $\alpha=1\times 10^{-4}$) to minimize the weighted excess loss.
\end{enumerate}
This iterative process continues for a fixed number of steps (100k steps), allowing the proxy model to converge to an optimal mixture $\bm{p}^*$ that maximizes the learning gain across all domains. The complete procedure is summarized in Algorithm~\ref{alg:promix}.

\begin{algorithm}[t]
\caption{ProMix: Proxy-Reference Weight Learning}
\label{alg:promix}
\begin{algorithmic}[1]
\State Train reference model $\bm{\theta}_R$ on $\mathcal{D}$ with a uniform mixture; freeze it.
\State Initialize proxy model $\bm{\theta}_P$; set weights $\bm{p}$ uniform over $\Delta$.
\For{$t = 1, \ldots, K$}
    \State Sample one minibatch $\mathcal{B}_i$ independently from each domain $\mathcal{D}_i$.
    \State Compute $\ell_i^P=L_{\mathcal{D}_i}(\bm{\theta}_P;\mathcal{B}_i)$ and $\ell_i^R=L_{\mathcal{D}_i}(\bm{\theta}_R;\mathcal{B}_i)$, detaching $\ell_i^R$ from the gradient graph.
    \State Update proxy parameters by descending the weighted excess loss:
    $\bm{\theta}_P \leftarrow \bm{\theta}_P - \alpha \nabla_{\bm{\theta}_P} \sum_{i} p_i (\ell_i^P-\ell_i^R)$
    \State Update weights by exponentiated gradient ascent (Eq.~\eqref{eq:weight_update}).
\EndFor
\State \Return $\bm{p}^*$: weights averaged over the final 10{,}000 steps.
\end{algorithmic}
\end{algorithm}

\subsection{Learning Data Mixture Weights}
\label{subsec:learning_weights}

Instead of directly minimizing the absolute loss on each domain, our proxy model learns to minimize the \textbf{excess loss} relative to the reference model. This is designed to prevent the model from becoming overly pessimistic and to yield a robust solution. The optimization objective for the proxy model is formalized as:
\begin{equation}
\min_{\bm{\theta}_P} \max_{\bm{p} \in \Delta} \sum_{i=1}^{N} p_i \cdot \left[ L_{\mathcal{D}_i}(\bm{\theta}_P) - L_{\mathcal{D}_i}(\bm{\theta}_R) \right] - \lambda \sum_{i=1}^{N} p_i \log p_i
\label{eq:proxy_objective}
\end{equation}
where the term in brackets represents the excess loss on domain $\mathcal{D}_i$. The second term equals $\lambda H(\bm p)$ with $H(\bm p)=-\sum_i p_i\log p_i$; since the inner problem \emph{maximizes} over $\bm p$, adding $+\lambda H(\bm p)$ favors high-entropy mixtures and discourages collapse (cf.\ Remark~\ref{rem:entropy}).

This minimax problem is solved using an iterative two-step optimization process:
\begin{enumerate}
    \item \textbf{Proxy Model Update}: For a fixed set of weights $\bm{p}$, we update the proxy model parameters $\bm{\theta}_P$ via gradient descent to minimize the weighted loss:
    \begin{equation}
    \bm{\theta}_P \leftarrow \bm{\theta}_P - \alpha \nabla_{\bm{\theta}_P} \left[ \sum_{i=1}^{N} p_i \cdot L_{\mathcal{D}_i}(\bm{\theta}_P) \right]
    \label{eq:proxy_update}
    \end{equation}
    where $\alpha$ is the learning rate for the model parameters (optimized with AdamW; Section~\ref{subsec:setup}).
    \item \textbf{Weight Update}: For a fixed proxy model $\bm{\theta}_P$, we update the weights $p_i$ via exponentiated-gradient ascent. With $g_i=L_{\mathcal{D}_i}(\bm{\theta}_P)-L_{\mathcal{D}_i}(\bm{\theta}_R)$, the entropy-regularized update is:
    \begin{equation}
    p_i \leftarrow \frac{p_i\exp\{\eta[g_i-\lambda(1+\log p_i)]\}}{\sum_{j=1}^{N}p_j\exp\{\eta[g_j-\lambda(1+\log p_j)]\}}
    \label{eq:weight_update}
    \end{equation}
    where $\eta$ is the learning rate for the weights. This update rule naturally maintains the simplex constraint $\bm{p} \in \Delta$.
\end{enumerate}
Individual excess losses $g_i$ may be negative when the proxy surpasses the reference on a domain; the simplex-constrained update in Eq.~\eqref{eq:weight_update} remains well defined in this regime, and the temporal averaging of $\bm p^*$ over the final 10{,}000 steps suppresses minibatch noise in the estimates of $g_i$.

\begin{remark}[Entropy regularization bounds worst-case concentration]
\label{rem:entropy}
For fixed excess losses $\bm g$, the inner objective in Eq.~\eqref{eq:proxy_objective} is strictly concave in $\bm p$, and its unique maximizer is $\bm p \propto \exp(\bm g/\lambda)$. Consequently $p_i/p_j = \exp\{(g_i-g_j)/\lambda\}$ for any two domains, and if the excess losses have range $G=\max_i g_i-\min_i g_i$, every domain retains weight at least $p_i \ge \frac{1}{N}e^{-G/\lambda}$.
As $\lambda \to 0$ this floor vanishes and the iterate collapses onto the domains with the largest excess; a positive $\lambda$ keeps every domain in the support with a floor that decays only exponentially in the excess range. Entropy regularization thus bounds the \emph{degree} of concentration, while the reference model determines its \emph{direction}; this division of labor matches the two ablations in Section~\ref{subsec:results}.
\end{remark}
This process is repeated for a set number of steps, and the reported weights $\bm{p}^*$ are the temporal average of the iterates over the final 10{,}000 steps after warm-up. In all experiments we use $\lambda = 0.1$, selected by a grid search over $\{0.01, 0.05, 0.1, 0.5, 1.0\}$ on validation loss (Section~\ref{subsec:setup}).

\subsection{Final Model Training}
\label{subsec:final_training}

Once the optimal weights $\bm{p}^*$ are obtained, they are frozen and used to guide the training of the final, large-scale VLA model $\bm{\theta}_L$. The training objective is:
\begin{equation}
\min_{\bm{\theta}_L} \sum_{i=1}^{N} p_i^* \cdot L_{\mathcal{D}_i}(\bm{\theta}_L)
\label{eq:final_training}
\end{equation}
This stage is highly efficient because the resource-intensive weight search never touches the large model. Whereas Re-Mix \cite{remix} must train a reference model and a robust proxy on the full corpus at (near-)final policy scale before training the final policy---two additional full passes over the data---ProMix performs the entire search with a 15M proxy against a frozen 15M reference: reference training (3 GPU-hours) plus proxy search (8 GPU-hours), i.e., 11 GPU-hours in total, measured with matched 100k proxy steps and with final-policy training excluded. Under the same conventions, a full-scale search costs twice the corresponding final-model budget---16 GPU-hours at the 50M comparison scale and 192 GPU-hours at the 7B scale---so the proxy search decouples mixture-search cost from the backbone scale. The complete accounting, including the 192-GPU-hour cost of a full-scale Re-Mix search under the same conventions, is given in Section~\ref{subsec:setup}.

We further validate this ``proxy-to-final'' transferability in Section \ref{sec:exp} by showing that $\bm{p}^*$ learned by a 15M model consistently outperforms uniform baselines on 50M and 7B architectures, without any re-tuning.

\section{Experiments}
\label{sec:exp}
Our experiments answer four questions: (1) Does \textbf{ProMix} outperform existing data mixture strategies in robotic imitation learning? (2) Is the ``proxy-reference'' architecture a crucial component of its success? (3) Do the learned mixture weights transfer effectively to larger-scale models? (4) Which domains benefit most from optimized mixing?

\subsection{Experimental Setup}
\label{subsec:setup}

We conducted our experiments on a diverse set of real-world robotics data to ensure that our findings are robust and generalizable.

\subsubsection{Datasets}

We use a subset of the \textbf{Open-X Embodiment Dataset \cite{openx202}}, a large-scale collection of robot manipulation data. We partition this dataset into distinct domains based on two primary criteria:
\begin{enumerate}
    \item \textbf{Robot Embodiment}: Data is grouped by the robot platform (e.g., WidowX, UR5, Franka). This allows us to test our method's ability to learn from varying robot kinematics and dynamics.
    \item \textbf{Task Category}: The training data is grouped by task type (e.g., pick-and-place, stacking, door opening). This tests our method's ability to balance training across different skill sets.
\end{enumerate}
The total corpus is 9500 episodes with an 80/20 train/validation split per domain; Table~\ref{tab:domain_stats} reports per-domain statistics.

Table~\ref{tab:domain_stats} specifies the domain partition used throughout the experiments. The five named task columns in Table~\ref{tab:main_results} are representative reporting groups; aggregate metrics are computed over all seven domains listed in Table~\ref{tab:domain_stats}.

\begin{table*}[!t]
\centering
\caption{Success Rate (\%) Comparison across Data Mixture Strategies on Open-X Embodiment at the 50M scale.}
\label{tab:main_results}
\begin{tabular}{l c c c c c c c}
\toprule
\multirow{2}{*}{Method} & \multirow{2}{*}{Avg.} & \multicolumn{5}{c}{Per-Domain Success Rate (\%)} & \multirow{2}{*}{Min Domain} \\
\cmidrule(lr){3-7}
& & P\&P & Stack. & Door & Drawer & Insert. & \\
\midrule
Uniform Mixing      & 68.4 $\pm$ 1.2 & 72.1 & 65.8 & 58.3 & 70.2 & 51.4 & 51.4 \\
Proportional Sampling & 64.1 $\pm$ 1.5 & 69.3 & 62.4 & 53.7 & 66.8 & 47.2 & 47.2 \\
Expert Weights       & 70.3 $\pm$ 1.3 & 74.5 & 68.1 & 61.2 & 72.4 & 54.6 & 54.6 \\
Re-Mix \cite{remix} & 74.8 $\pm$ 0.9 & 77.2 & 72.5 & 71.8 & 76.3 & 65.4 & 65.4 \\
\midrule
\textbf{ProMix (Ours)} & \textbf{76.2 $\pm$ 0.7} & \textbf{78.4} & \textbf{76.1} & \textbf{78.8} & \textbf{78.0} & \textbf{72.3} & \textbf{68.5} \\
\bottomrule
\end{tabular}
\vspace{1mm}
\parbox{\linewidth}{\footnotesize \textit{Note:} ``Min Domain'' is the worst-performing domain over \emph{all} seven evaluation domains, including the two not displayed above.}
\vspace{-2mm}
\end{table*}

\begin{table*}[!t]
\centering
\caption{Ablation Study: Role of the Reference Model and Entropy Regularization. Variant weights are evaluated by training the 50M final model as in Table~\ref{tab:main_results}; compute follows the accounting conventions described in Section~\ref{subsec:setup}. ``Weight Entropy'' is $H(\bm p^*)=-\sum_i p_i^*\log p_i^*$ (lower values indicate more concentrated weights).}
\label{tab:ablation}
\begin{tabular}{l c c c c}
\toprule
Method & Avg. SR (\%) & Min Domain SR (\%) & Weight Entropy & Proxy Steps \\
\midrule
ProMix w/o Reference & 72.4 $\pm$ 1.1 & 54.2 & 0.82 & 150k \\
ProMix w/o Entropy Reg. & 73.8 $\pm$ 1.4 & 48.7 & 0.31 & 100k \\
\textbf{ProMix (Full)} & \textbf{76.2 $\pm$ 0.7} & \textbf{68.5} & \textbf{1.24} & 100k \\
\bottomrule
\end{tabular}
\vspace{1mm}
\parbox{\linewidth}{\footnotesize \textit{Note:} w/o Entropy Reg.\ shares Full's cost and step budget; only the resulting weight distribution differs.}
\vspace{-2mm}
\end{table*}

\begin{table}[!t]
\centering
\caption{Domain statistics for the Open-X subset used in our experiments. Episode counts are the number of trajectories whose language annotation matches the task domain in the source Open-X dataset. ``Split'' reports the native train/test division of each source dataset.}
\label{tab:domain_stats}
\scriptsize
\resizebox{\columnwidth}{!}{%
\begin{tabular}{l l l r c}
\toprule
Domain & Embodiment & Task type & Episodes & Split \\
\midrule
Pick-and-place   & Kinova Jaco& pick/place & 1{,}085 & 80/20 \\
Stacking cups    & Google Robot   & stack   & 368     & 80/20 \\
Door opening     & Hello Robot Stretch & door  & 484     & 80/20 \\
Drawer opening   & WidowX 250     & drawer     & 774     & 80/20 \\
Peg insertion    & Franka Emika Panda & insert & 631  & 80/20 \\
Table wiping     & Google Robot   & wipe     & 4{,}786 & 80/20 \\
Mug manipulation & Franka Emika Panda     & mug    & 1385      & 80/20 \\
\bottomrule
\end{tabular}
}%
\end{table}

\subsubsection{Evaluation Protocol}

We evaluate the trained policies using \textbf{closed-loop real-robot rollouts} on \textbf{the corresponding physical robot platform for each domain}. The evaluation uses \textbf{the released OXE evaluation splits} across \textbf{15} held-out tasks and \textbf{20} trials per task, following the task definitions in the released benchmark split. A trial succeeds when \textbf{the policy completes the task within 30 seconds without exceeding joint limits or colliding with the environment}. We report the mean success rate over domains and average the resulting values over 5 random seeds; confidence intervals are 95\% confidence intervals computed over the seed-level results.

\subsubsection{Model Architectures}

Reference, proxy, and the 15M/50M final models share the compact VLA architecture of Section~\ref{subsubsec:architecture}, trained with a maximum sequence length of 8 actions. Reference and proxy models each have approximately 15M trainable parameters, excluding frozen encoders. We validate transferability of the final model ($\theta_L$) at three scales: 15M (the proxy scale), 50M (medium), and 7B (large). The 7B experiment uses \textbf{the released OpenVLA 7B checkpoint} \cite{kim2025openvla}, with \textbf{frozen visual and language backbones and trainable decoder layers}; Uniform and ProMix use the same initialization, optimizer, data budget, and evaluation protocol. OpenVLA uses a different pretrained VLA architecture, so the 7B result measures transfer across both scale and architecture; it is not a strict width-only scaling experiment.

\subsubsection{Training Details}

All models are trained using the AdamW optimizer with weight decay $0.01$. The learning rate for model parameters ($\alpha$) is set to $10^{-4}$ with cosine decay, and the learning rate for the domain weights ($\eta$) is set to $10^{-2}$. The entropy coefficient is $\lambda=0.1$, selected using \textbf{a grid search over $\{0.01, 0.05, 0.1, 0.5, 1.0\}$ on validation loss}. Proxy and reference models train for 100{,}000 steps (warmup 1{,}000); the final models train for 300k (15M), 1M (50M), and 50k (7B) steps. The mini-batch size is set to 128. We report mean $\pm$ 95\% CI over 5 seeds. ProMix's improvement over Uniform Mixing is large relative to these intervals; its $+1.4\%$ difference from Re-Mix, however, lies within the overlap of the two methods' 95\% CIs, so we regard ProMix as statistically comparable to Re-Mix in average success rate, with its advantage being the mixture-search cost.

For the mixture-search comparison, the 11 GPU-hours for ProMix comprise \textbf{reference training (3h) and proxy training (8h)}; evaluation overhead, hyperparameter-selection runs (e.g., the $\lambda$ grid), and final-policy training are excluded from all reported figures. Although the compact models have only approximately 15M trainable parameters, each update still executes frozen encoders and per-domain loss evaluation; the accelerators are therefore an execution choice for throughput measurement, not a minimum hardware requirement. A full-scale Re-Mix search under the same conventions requires a reference model and a robust model, each trained like a final policy, and therefore costs twice the final-model budget: 16 GPU-hours at the 50M comparison scale and 192 GPU-hours at the 7B scale. The final models themselves cost 2.1, 8.0, and 96.0 GPU-hours at the 15M, 50M, and 7B scales (on 1, 4, and 8 H100 GPUs respectively). Because the ProMix search cost is fixed at 11 GPU-hours while the full-scale search cost grows with the backbone, we report GPU-hours at matched model scales and make no hardware-normalized claims; ablation variants use the same convention, with step budgets in Table~\ref{tab:ablation}.

\subsection{Results and Analysis}
\label{subsec:results}

\subsubsection{Main Results: Comparison with Baselines}

We compare \textbf{ProMix} against uniform mixing, proportional sampling (domains sampled by size), a hand-tuned expert weight vector, and a reimplementation of Re-Mix \cite{remix}, the state-of-the-art data mixture optimizer for robotics.

Table~\ref{tab:main_results} summarizes the performance of our method against these baselines. \textbf{ProMix} achieves an average success rate of $76.2\%$, improving over uniform mixing by $+7.8\%$ and over Re-Mix by $+1.4\%$. Because the 95\% CIs of ProMix and Re-Mix overlap, we read this residual gap as the two methods being comparable at this scale, with ProMix's advantage being an order-of-magnitude cheaper search (Fig.~\ref{fig:results_summary}(a)); since ProMix differs from Re-Mix in both the search scale and the entropy regularizer, the gap should not be attributed to either factor alone, and Table~\ref{tab:ablation} isolates each component relative to ProMix. Notably, the minimum domain success rate improves from $51.4\%$ (uniform) to $68.5\%$ (+17.1\%), validating the minimax formulation's ability to lift challenging, low-resource domains. Proportional sampling underperforms uniform mixing, indicating that larger domains dominate training when naively weighted by size.

\subsubsection{Ablation Studies}

To demonstrate the critical role of the ``reference model'' in our framework, we conduct an ablation study by removing this component. In this variant, the proxy model is trained to directly minimize the worst-case absolute loss---i.e., standard group DRO, obtained by removing the excess-loss baseline from Eq.~\eqref{eq:proxy_objective}. We also evaluate a variant without entropy regularization ($\lambda = 0$) to assess its role in preventing weight collapse. All variants use the same 15M proxy architecture and hardware as the full model, and each variant's learned weights are evaluated by training the 50M final model under the protocol of Table~\ref{tab:main_results}.

Table~\ref{tab:ablation} reports the ablation results. Removing the reference model causes a $3.8\%$ drop in average success rate and a substantial $14.3\%$ drop in minimum domain performance, consistent with the excess-loss formulation in Eq.~\eqref{eq:proxy_objective} being essential for robust optimization: as illustrated in Remark~\ref{rem:toy}, without the reference baseline the DRO proxy over-weights domains with the highest \emph{absolute} loss---typically the noisiest ones---rather than those with genuine headroom. In our runs this variant was allotted $50\%$ more proxy steps (150k vs.\ 100k), as the absence of the reference baseline makes the excess-loss signal noisier; even with this larger budget it remains below the full method. Removing entropy regularization leads to weight collapse: the weight entropy drops from $1.24$ to $0.31$, and the minimum domain success rate falls from $68.5\%$ to $48.7\%$. This indicates that the regularizer prevents over-concentration on a few domains (Remark~\ref{rem:entropy}) without adding training cost---the variant shares the per-step cost and step budget of the full method. The difference lies entirely in the quality of the resulting weight distribution.

\begin{figure}[!b]
\centering
\includegraphics[width=0.97\columnwidth]{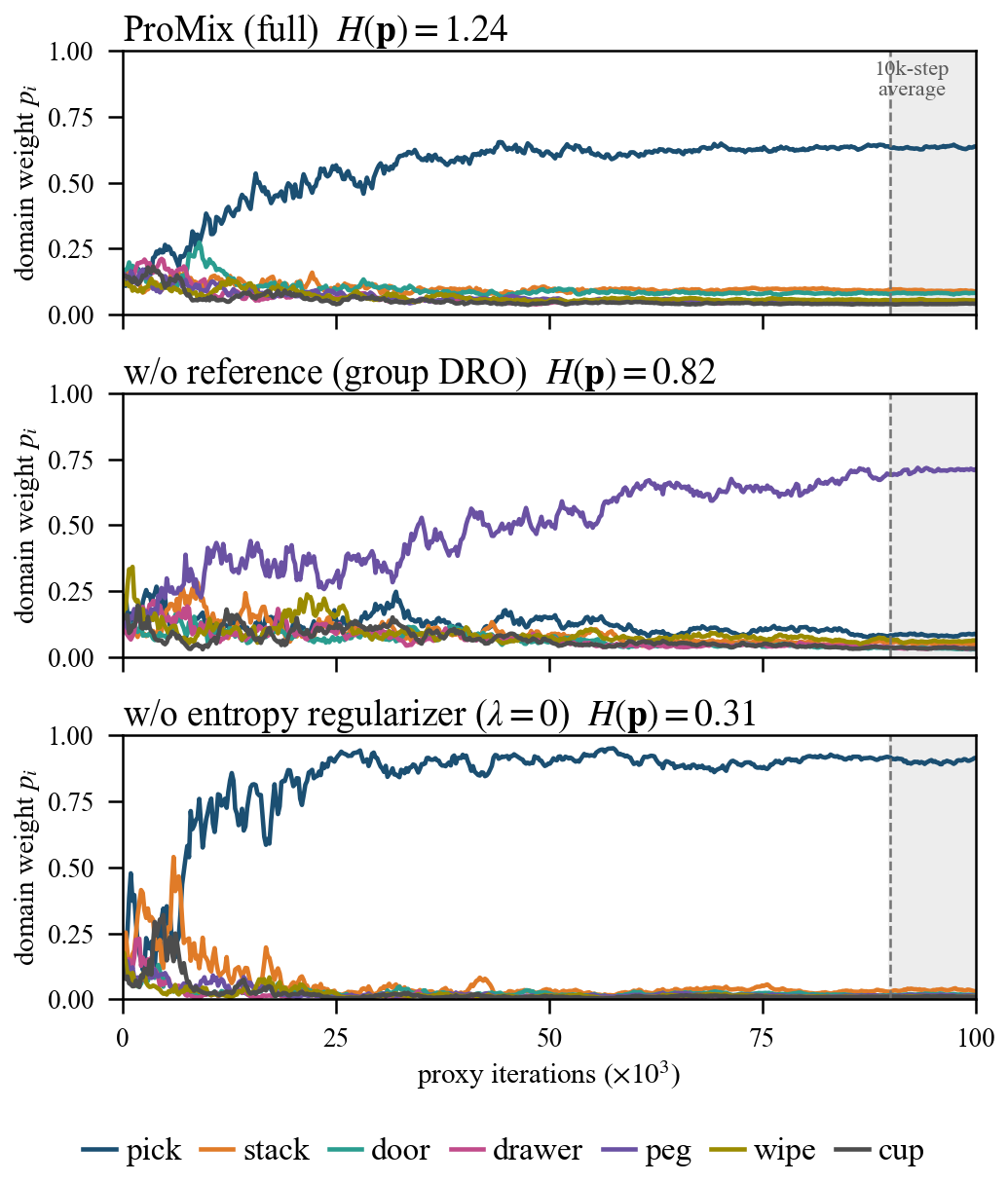}
\caption{\textbf{Evolution of the domain weights during the proxy search.} Each panel traces the seven domain weights $p_i$ from uniform initialization (top to bottom: full method, without the reference model, without entropy regularization); the shaded band marks the final 10k-step window averaged into $\bm p^{*}$.}
\label{fig:weights}
\end{figure}

Fig.~\ref{fig:weights} shows how the two ablations reshape the mixture over training. Under the full method the weights stabilize within roughly 40k iterations, so the 10k-step average is taken over a converged distribution. The reference-free variant instead keeps concentrating mass on one domain, starving the rest of the gradient; removing the regularizer collapses the weights onto one domain within roughly 25k iterations. This supports the division of labor suggested by Remark~\ref{rem:entropy}: the reference model determines \emph{which} domains attract mass, while the regularizer determines \emph{how} concentrated it becomes.

\subsubsection{Cross-Scale Generalization}

Following scaling laws, we fix and transfer $\mathbf{p}^*$ from the 15M proxy to 50M and 7B models without extra tuning. This isolates the effect of the learned distribution from model capacity.

\begin{figure*}[!t]
\centering
\includegraphics[width=0.85\textwidth]{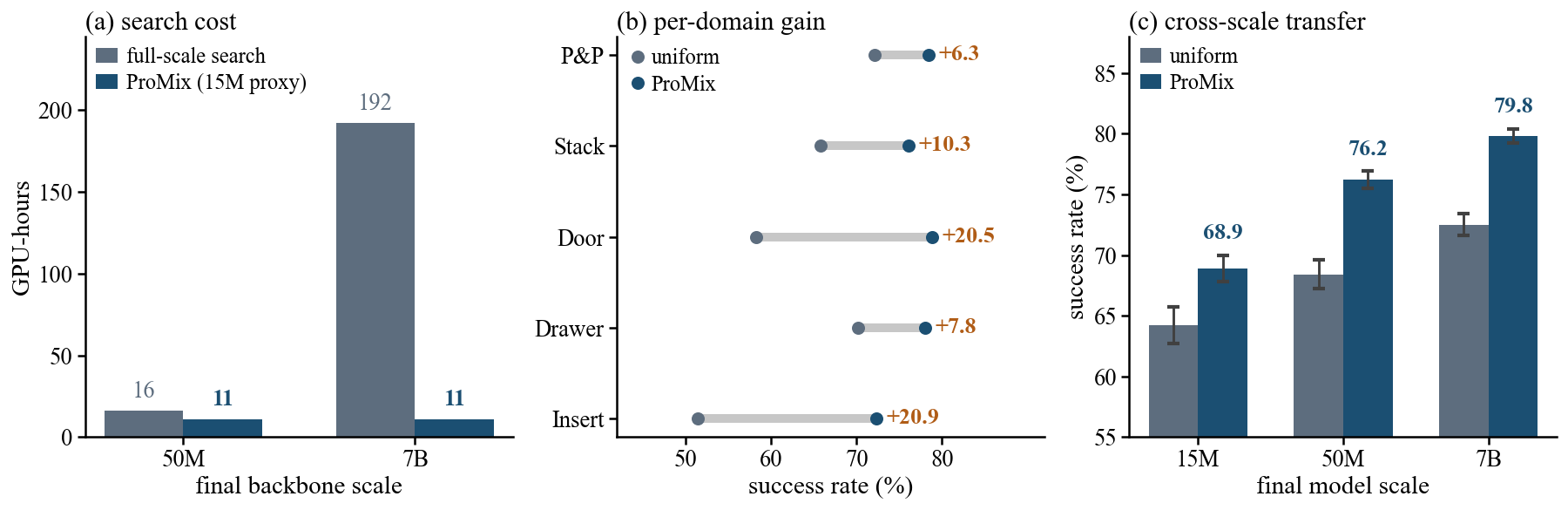}
\caption{\textbf{Summary of the main empirical effects.} (a) Mixture-search cost: a full-scale search costs 16 GPU-hours at the 50M comparison scale and 192 GPU-hours at 7B, whereas the ProMix search is fixed at 11 GPU-hours regardless of backbone. (b) Per-domain improvement over uniform mixing at the 50M scale. (c) Transfer of the 15M-learned mixture to final models at three scales without re-tuning (95\% CIs over 5 seeds).}
\label{fig:results_summary}
\end{figure*}

Fig.~\ref{fig:results_summary}(c) shows consistent gains over uniform mixing across all scales ($+7.3\%$ to $+11.4\%$ relative), confirming that \textbf{ProMix} enables a \textit{compute once, deploy anywhere} workflow: a mixture search that fits on a single GPU yields benefits that persist---rather than vanish---as model scale grows.

\section{Conclusion}
\label{sec:conclusion}

In this work, we presented \textbf{ProMix}, a compute-efficient framework for optimizing data mixtures in large-scale robotic imitation learning. ProMix learns domain weights with a lightweight proxy-reference pair---at a mixture-search cost of 11 GPU-hours that is fixed regardless of backbone scale, versus 16--192 GPU-hours for full-scale searches at 50M--7B under matched accounting---and transfers the learned weights unchanged to 50M and 7B VLA backbones. Beyond efficiency, our controlled ablations isolate \emph{why} the proxy-reference design matters in robotics: without the reference model, DRO over-weights noisy domains, and without entropy regularization, the weights collapse onto a few sources. Experiments on 9500 real-world episodes from the Open-X Embodiment Dataset show that ProMix lifts the average success rate from 68.4\% to 76.2\% and the worst-domain rate from 51.4\% to 68.5\%, making advanced data optimization accessible to researchers with limited resources.

\textbf{Limitations and future work.} ProMix operates at the domain level: it re-weights whole data sources but does not filter individual low-quality trajectories, so final performance remains bounded by corpus quality; combining domain mixing with demonstration-level curation \cite{hejna2025deminf} is a promising direction. Our domains are defined by robot embodiment and task category; semantically defined domains (skills, scenes, instruction families) may yield different optimal mixtures. The excess-loss objective also presumes the frozen reference model is competent enough on every domain to provide a meaningful baseline; when the reference fails catastrophically, the entropy regularizer mitigates but does not eliminate weight collapse. Extending the framework to pruning and continual pre-training, and validating learned mixtures across more embodiments, remain future work.


\bibliographystyle{IEEEtran}
\bibliography{refs}

\end{document}